\documentclass[a4paper]{article}

\usepackage{INTERSPEECH2018}
\usepackage{subfigure}
\usepackage{stfloats}

\title{AISHELL-2: Transforming Mandarin ASR Research Into Industrial Scale}
\name{Jiayu Du$^1$, Xingyu Na$^1$, Xuechen Liu$^1$, Hui Bu$^2$}
\address{
  $^1$AISHELL foundation$^*$\thanks{* AISHELL foundation is a non-profit online organization, dedicated to pushing forward speech industry via open-sourcing database to research institutes and contributing codes to open-source speech community.} \\
  $^2$Beijing Shell Shell Technology Co. Ltd., Beijing, China}
\email{aishell.foundation@gmail.com, buhui@aishelldata.com}

\begin{document}

\maketitle
\begin{abstract}
AISHELL-1 is by far the largest open-source speech corpus available for Mandarin
speech recognition research. It was released with a baseline system containing solid
training and testing pipelines for Mandarin ASR. In AISHELL-2, 1000 hours of
clean read-speech data from iOS is published, which is free for academic usage. On top of
AISHELL-2 corpus, an improved recipe is developed and released, containing key
components for industrial applications, such as Chinese word segmentation,
flexible vocabulary expension and phone set transformation etc.
Pipelines support various state-of-the-art techniques, such
as time-delayed neural networks and Lattic-Free MMI objective funciton.
In addition, we also release dev and test data from other channels(Android and Mic).
For research community, we hope that AISHELL-2
corpus can be a solid resource for topics like transfer learning and robust
ASR. For industry, we hope AISHELL-2 recipe can be a helpful reference for
building meaningful industrial systems and products.
\end{abstract}
\noindent\textbf{Index Terms}: Speech recognition, Mandarin ASR, Industrial Speech Recognition

\section{Introduction}

Automatic Speech Recognition (ASR) is a major application domain in the bloom of Artificial
Intelligence (AI). Huge effort has been made from both research community and industry to improve ASR system performance. Among all solutions proposed, deep learning approach has been dominating for the last half decade. Given enough data, neural network (NN) models generally perform better in terms of recognition accuracy, and turn out to be more robust.
From industrial perspective, accessing and collecting large amount of speech data has become easier than ever before, with emerging market of smart phones and various other smart devices.
However, on the other hand, research community still has limited-access to real-world application data. As a result, improvements in research community do not always scale well to industrial scenarios.
In computer vision, there are many high quality free data sets which transform research efforts into industrial applications, such as ImageNet~\cite{imagenet} and COCO~\cite{coco}.
In Mandarin ASR, although there are corpus like thchs30~\cite{thchs30} and hkust~\cite{hkust1}, a large-scale high-quality free corpus is still needed.

In AISHELL-1~\cite{aishell1}, we released 170 hours of Mandarin speech with high quality human transcriptions.
Various training and evaluation recipes based on such corpus have been developed in Kaldi~\cite{kaldi}, which is a robust and widely-acknowledged framework for speech research.
To the best knowledge of the authors, it is the first fully open-sourced system for Mandarin ASR with high quality Mandarin speech data(e.g. ~\cite{do2017, do2018_1, do2018_2}).

Furthering the success from AISHELL-1 with efforts, in this paper, we introduce AISHELL-2, an open-sourced, self-contained baseline for industrial-scale Mandarin ASR research.
On one hand, 1000 hours of iPhone-recorded speech data is released. On the other hand, baseline recipes(containing must-have components such as Chinese word segmentation, customizable Chinese lexicon etc) are published into Kaldi, following what have been done with AISHELL-1 formally.
Moreover, besides training data(iOS), we release development and test data from 3 acoustic channels(iOS, Android and Mic).
We hope these resources would be helpful to transform Mandarin ASR research into industrial scale.

The rest of this paper is organized as follows: The details of AISHELL-2 corpus data is introduced in Section 2.
Section 3 describes the Mandarin ASR pipeline for producing baseline system and Section 4
presents system performance on test sets from different acoustic channels.

\section{AISHELL-2 corpus}

\begin{figure}[h]
  \centering
  \includegraphics[width=\linewidth]{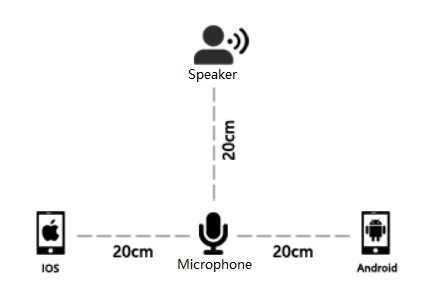}
  \caption{Recording setup}
  \label{fig:setup}
\end{figure}

AISHELL-2 corpus contains 1000 hours of clean reading-speech data, which will be released to academic research community for free. Raw data was recorded via three parallel acoustic channels - a high fidelity microphone (Mic), an Android smartphone (Android) and an iPhone (iOS). The relative position between speaker and devices are shown in Figure~\ref{fig:setup}. Data from iPhone, i.e. the iOS channel is open-sourced. Speaker, environment and content coverage are explained as below:
\begin{itemize}
\item \textbf{Speaker information.} There are 1991 speakers participated in the recording, including 845 male and 1146 female. Age of speaker ranges from 11 to over 40. Ideally the speakers shall speak everything to be recorded in Mandarin, while there are some slight accent variations. Generally speaking of the accents, there were 1293 speakers using Northern ones, 678 speakers using Southern ones and 20 speakers use other accents during recording.
\item \textbf{Recording environment.} 1347 speakers are recorded in a studio, while the rest are in a living room with natural reverberation.
\item \textbf{Content of speech}. The content of the recording covers 8 major topics: voice commands such as IoT device control and digital sequential input, places of interest, entertainment, finance, technology, sports, English spellings and free speaking without specific topic. The total number of prompts is around half a million.
\end{itemize}

\noindent Aside from AISHELL-2 corpus introduced above, which is supposed to be used for training,
we also provide development and test sets.
Development set contains 2500 utterances from 5 speakers and test set contains 5000 utterances from 10 speakers.
Each speaker contributed approximately half an hour of speech, covering 500 prompts.
First 7 prompts of all speakers are extracted from high frequency online queries. Speaker-gender is balanced as well.

\section{AISHELL-2 recipe}

As briefly mentioned in Section 1, based on AISHELL-2 corpus, several recipes are released to Kaldi repository as a complete collection, including data and lexicon preparation, language model training, Gaussian mixture model (GMM) and Neural Network training, as well as test set evaluation procedure.

\subsection{Lexicon and word segmentation}

\begin{figure}[t]
  \centering
  \includegraphics[width=0.8\linewidth]{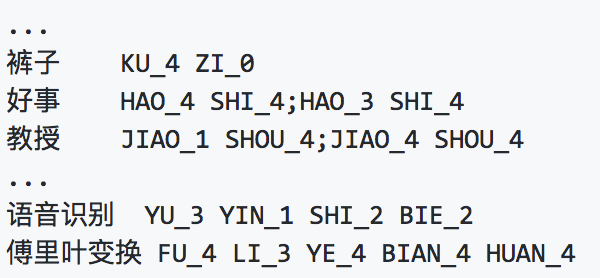}
  \caption{Layer 1 of DaCiDian}
  \label{fig:lex1}
\end{figure}

\begin{figure}[t]
  \centering
  \includegraphics[width=0.8\linewidth]{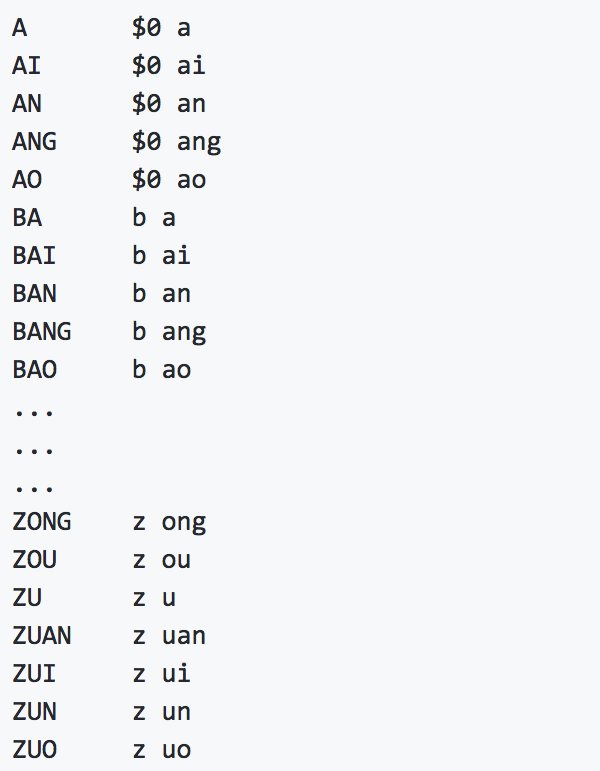}
  \caption{Layer 2 of DaCiDian}
  \label{fig:lex2}
\end{figure}

Unlike English ASR system, Mandarin ASR often requires sophisticated word segmentation.
In AISHELL-1, word segmentation was implemented via forward
maximum matching algorithm, based on a variation of open-source mandarin
dictionary CC-CEDIT\footnote{https://cc-cedict.org/wiki}. In AISHELL-2, an
open-source Chinese dictionary called DaCiDian is
released\footnote{https://github.com/aishell-foundation/DaCiDian}. In most
common Chinese dictionaries, words are directly mapped to phonemes. While in
DaCiDian, this mapping is decomposed into 2 independent layers. An exemplar
DaCiDian structure is shown in Figure~\ref{fig:lex1} and Figure~\ref{fig:lex2}. The first layer maps word to PinYin syllables~\cite{pinyin}. Anyone who is
  familiar with PinYin (basically every Mandarin speaker) can enrich DaCiDian's
  vocabulary by adding new words into this layer. The second layer is a mapping from Pinyin syllable to phoneme. ASR system
  developers can easily adapt DaCiDian to their own phone set by redefining this
  layer of mapping.


In terms of word segmentation, we choosed a popular and easy-to-use open-source
toolkit called Jieba~\cite{jieba}, it implemented a trie-tree based algorithm and
supports vocabulary customization. Based on DaCiDian and Jieba, we provide a
script to segment AISHELL-2 transcription and language model text.

\subsection{Acoustic model}

Acoustic training contains two stages: GMM-HMM state model training based on maximum likelihood and later the training of a hybrid DNN-HMM state estimator. Both can be implemented by calling standard methods and designing corresponding recipes in Kaldi toolkit.

The GMM models were firstly trained using 13 dimensional MFCC plus pitches, which made the input dimension 16.
A monophone model was trained to set a starting point for the triphone models. A small triphone model and a larger triphone model were then consecutively trained using delta features. After that, a more sophisticated feature transform method was applied to replace the delta features. Linear discriminant analysis (LDA) was applied on stack of frames to reduce the dimension and MLLT-based global transform is estimated iteratively. This follows a standard setup pipeline in the majority of available Kaldi recipes. The resulting number of physical GMM states from the four steps were 605, 3216, 5720 and 8080 respectively.

GMM training of the AISHELL-2 stopped at the speaker independent stage, without speaker dependent transform involved, such as fMLLR. For industrial-scale corpus, it is not worth spending too much time and computation power at GMM-HMM training stage, since the final system performance primarily depends on later neural network models. Therefore, we adopted speaker independent GMM training and pushed the speaker dependent steps to later DNN training phase, as described below.

Based on the tied-triphone state alignments from GMM, a time-delayed neural network (TDNN, \cite{tdnn}) is then configured and trained. It has 8 layers in total, with 1280 hidden units in each layer. It is, as described in \cite{tdnn}, not fully connected - the input of each hidden layer is a frame-wise spliced output of its preceding layer. The input feature was high-resolution MFCC with cepstral normalization plus pitches, which made its dimension 43. Note that for each frame-wise input a 100-dimensional i-vector~\cite{ivector} was also attached, whose extractor was trained based on the corpus itself. The corresponding diagonal universal background model (UBM) was trained using a quarter of training features. This indicates that different from GMM-level training, we encoded speaker information here to produce a stronger baseline for research community. The network was trained using lattice-free maximum mutual information (LFMMI, \cite{lfmmi}) as the objective function. More configurations about NN training itself such as lattice generation and tree topology can be found at \cite{lfmmi} and AISHELL-2 exemplar scripts in Kaldi repository \footnote{https://github.com/kaldi-asr/kaldi/egs/aishell2}.

\subsection{Language model}

A trigram language model was trained on 5.7 million-word speech transcripts from AISHELL-2. Out-of-vocabulary (OOV) words were mapped as $<$UNK$>$. The 3-order ARPA language model is trained using Kneser-Ney smoothing, with 516552 unigrams, 1498603 bigrams and 932475 trigrams, respectively.

\section{Experiment and evaluations}

\begin{table*}[tp]
  \caption{Baseline system results and training time}
  \label{tab:base}
  \centering
  \begin{tabular}{ llllllll }
    \toprule
    CER               &  dev\_android           &  dev\_ios           &  dev\_mic           & test\_android            &  test\_ios           &  test\_mic          &  Training time in hours         \\
    \midrule
    Mono        &  47.08                 &  43.37             &  47.33             &  45.40                  &  44.81              &  44.28             &  0.5                   \\
    tri1    &  26.61                 &  22.94             &  26.55             &  26.08                  &  24.79              &  25.36             &  1                     \\
    tri2    &  24.59                 &  21.47             &  24.59             &  23.82                 &  22.69              &  23.37             &  2                     \\
    tri3(LDA+MLLT)          &  22.24                 &  18.86             &  22.47             &  21.00                  &  19.77              &  21.10             &  2.5                   \\
    Chain-TDNN              &  10.43                 &  9.10             &  11.84         &  9.59                  &  8.81             &  10.87              &  15                    \\
    \bottomrule
  \end{tabular}
\end{table*}

Presented baseline system was trained on a standalone server, with 36 cores of Intel Xeon (2.3GHz) for all cpu-based steps and 4
Tesla K80 processors for DNN model training. Character Error Rate (CER) was used as the evaluation metric.

Results along with training time of each stage, are presented in Table~\ref{tab:base}.
Note that during training, we only used open-sourced iOS data, while evaluated these models on dev and test sets for all 3 channels (Android, Mic, and iOS). System performance on iOS outperformed Android and Mic, which is as expected due to better acoustic channel condition matching.

\section{Conclusions}

In this paper, we generally introduce AISHELL-2, a 1000-hour Mandarin ASR corpus, freely available to research community. In the meantime, we present a self-contained recipe in Kaldi toolkit as a research baseline. We hope this open-source project provides essential ingredients for researchers to explore more scalable and practical solutions regarding to industrial scenarios for Mandarin speech recognition.

\section{Acknowledgements}

The authors would like to thank all other members of AISHELL foundation who contributed to this project and Emotech Labs who provided computational resources for producing most recent system performance statistics.

\bibliographystyle{IEEEtran}

\bibliography{mybib}

\begin{thebibliography}{10}
\providecommand{\url}[1]{#1}
\csname url@samestyle\endcsname
\providecommand{\newblock}{\relax}
\providecommand{\bibinfo}[2]{#2}
\providecommand{\BIBentrySTDinterwordspacing}{\spaceskip=0pt\relax}
\providecommand{\BIBentryALTinterwordstretchfactor}{4}
\providecommand{\BIBentryALTinterwordspacing}{\spaceskip=\fontdimen2\font plus
\BIBentryALTinterwordstretchfactor\fontdimen3\font minus
  \fontdimen4\font\relax}
\providecommand{\BIBforeignlanguage}[2]{{%
\expandafter\ifx\csname l@#1\endcsname\relax
\typeout{** WARNING: IEEEtran.bst: No hyphenation pattern has been}%
\typeout{** loaded for the language `#1'. Using the pattern for}%
\typeout{** the default language instead.}%
\else
\language=\csname l@#1\endcsname
\fi
#2}}
\providecommand{\BIBdecl}{\relax}
\BIBdecl

\bibitem{imagenet}
J.~Deng, W.~Dong, R.~Socher, L.-J. Li, K.~Li, and L.~Fei-Fei, ``{ImageNet: A
  Large-Scale Hierarchical Image Database},'' in \emph{CVPR09}, 2009.

\bibitem{coco}
T.~Lin, M.~Maire, S.~J. Belongie, L.~D. Bourdev, R.~B. Girshick, J.~Hays,
  P.~Perona, D.~Ramanan, P.~Doll{\'{a}}r, and C.~L. Zitnick, ``{Microsoft COCO:
  Common Objects in Context},'' \emph{CoRR}, vol. abs/1405.0312, 2014.

\bibitem{thchs30}
\BIBentryALTinterwordspacing
D.~Wang and X.~Zhang, ``{THCHS-30} : {A} free chinese speech corpus,''
  \emph{CoRR}, vol. abs/1512.01882, 2015. [Online]. Available:
  \url{http://arxiv.org/abs/1512.01882}
\BIBentrySTDinterwordspacing

\bibitem{hkust1}
\BIBentryALTinterwordspacing
P.~Fung, S.~Huang, and D.~Graff. (2005) Hkust mandarin telephone speech, part
  1. [Online]. Available: \url{https://catalog.ldc.upenn.edu/LDC2005S15}
\BIBentrySTDinterwordspacing

\bibitem{aishell1}
H.~Bu, J.~Du, X.~Na, B.~Wu, and H.~Zheng, ``{AIShell-1: An Open-Source Mandarin
  Speech Corpus and A Speech Recognition Baseline},'' in \emph{Oriental COCOSDA
  2017}, 2017, p. Submitted.

\bibitem{kaldi}
D.~Povey, A.~Ghoshal, G.~Boulianne, L.~Burget, O.~Glembek, N.~Goel,
  M.~Hannemann, P.~Motlicek, Y.~Qian, P.~Schwarz, J.~Silovsky, G.~Stemmer, and
  K.~Vesely, ``{The Kaldi Speech Recognition Toolkit},'' in \emph{IEEE 2011
  Workshop on Automatic Speech Recognition and Understanding}, 2011.

\bibitem{do2017}
V.~H. Do, N.~F. Chen, B.~P. Lim, and M.~A. Hasegawa-Johnson, ``{Multitask
  Learning for Phone Recognition of Underresourced Languages Using Mismatched
  Transcription},'' \emph{IEEE/ACM Transactions on Audio, Speech, and Language
  Processing}, vol.~26, no.~3, pp. 501--514, March 2018.

\bibitem{do2018_1}
\BIBentryALTinterwordspacing
J.~Li, X.~Wang, Y.~Zhao, and Y.~Li, ``Gated recurrent unit based acoustic
  modeling with future context,'' \emph{CoRR}, vol. abs/1805.07024, 2018.
  [Online]. Available: \url{http://arxiv.org/abs/1805.07024}
\BIBentrySTDinterwordspacing

\bibitem{do2018_2}
Y.~Zhang, P.~Zhang, and Y.~Yan, ``Data augmentation for language models via
  adversarial training,'' \emph{ACTA AUTOMATICA SINICA}, vol. 44.5, pp.
  891--900, 2018.

\bibitem{pinyin}
S.~Duanmu, \emph{{The phonology of standard Chinese}}.\hskip 1em plus 0.5em
  minus 0.4em\relax Oxford University Press, 2007.

\bibitem{jieba}
J.~Sun, ``{Jieba Chinese word segmentation tool},'' 2012.

\bibitem{tdnn}
V.~Peddinti, D.~Povey, and S.~Khudanpur, ``A time delay neural network
  architecture for efficient modeling of long temporal contexts,'' in
  \emph{Proceedings of Interspeech}.\hskip 1em plus 0.5em minus 0.4em\relax
  ISCA, 2015.

\bibitem{ivector}
N.~Dehak, P.~J. Kenny, R.~Dehak, P.~Dumouchel, and P.~Ouellet, ``Front-end
  factor analysis for speaker verification,'' \emph{IEEE Transactions on Audio,
  Speech, and Language Processing}, vol.~19, no.~4, pp. 788--798, May 2011.

\bibitem{lfmmi}
D.~Povey, V.~Peddinti, D.~Galvez, P.~Ghahremani, V.~Manohar, X.~Na, Y.~Wang,
  and S.~Khudanpur, ``Purely sequence-trained neural networks for asr based on
  lattice-free mmi,'' in \emph{Proceedings of Interspeech}.\hskip 1em plus
  0.5em minus 0.4em\relax ISCA, 2016, pp. 2751--2755.

\end{thebibliography}

% \begin{thebibliography}{9}
% \bibitem[1]{Davis80-COP}
%   S.\ B.\ Davis and P.\ Mermelstein,
%   ``Comparison of parametric representation for monosyllabic word recognition in continuously spoken sentences,''
%   \textit{IEEE Transactions on Acoustics, Speech and Signal Processing}, vol.~28, no.~4, pp.~357--366, 1980.
% \bibitem[2]{Rabiner89-ATO}
%   L.\ R.\ Rabiner,
%   ``A tutorial on hidden Markov models and selected applications in speech recognition,''
%   \textit{Proceedings of the IEEE}, vol.~77, no.~2, pp.~257-286, 1989.
% \bibitem[3]{Hastie09-TEO}
%   T.\ Hastie, R.\ Tibshirani, and J.\ Friedman,
%   \textit{The Elements of Statistical Learning -- Data Mining, Inference, and Prediction}.
%   New York: Springer, 2009.
% \bibitem[4]{YourName17-XXX}
%   F.\ Lastname1, F.\ Lastname2, and F.\ Lastname3,
%   ``Title of your INTERSPEECH 2018 publication,''
%   in \textit{Interspeech 2018 -- 19\textsuperscript{th} Annual Conference of the International Speech Communication Association, September 2-6, Hyderabad, India Proceedings, Proceedings}, 2018, pp.~100--104.
% \end{thebibliography}

\end{document}